\documentclass[10pt, a4paper]{article}

\usepackage[]{lrec2026} 

\usepackage{times}
\usepackage{latexsym}

\usepackage[T1]{fontenc}

\usepackage[utf8]{inputenc}

\usepackage{microtype}

\usepackage{inconsolata}

\usepackage{graphicx}
\usepackage{multirow}
\usepackage{comment}
\usepackage{subcaption}
\usepackage{caption}
\usepackage{booktabs}
\makeatletter
\let\@cite@ofmt\@firstofone
\def\@biblabel#1{}
\def\@cite#1#2{{#1\if@tempswa , #2\fi}}
\makeatother
\newlength{\cslhangindent}
\newlength{\csllabelwidth}
{\begin{list}{}{%
\setlength{\itemindent}{0pt} \setlength{\leftmargin}{0pt} \setlength{\parsep}{0pt}
\ifodd #1 \setlength{\leftmargin}{\cslhangindent} \setlength{\itemindent}{-1\cslhangindent} \fi
\setlength{\itemsep}{#2\baselineskip}}}
{\end{list}}
\usepackage{calc}

\title{Readability Measures and Automatic Text Simplification:\\ In the Search of a Construct}

\name{Rémi Cardon\textsuperscript{1}, A. Seza Do\u{g}ruöz\textsuperscript{2}}

\address{\textsuperscript{1}Department of Computer Science and Engineering, Universidad Carlos III de Madrid \\ \textsuperscript{2}LT3 / IDLab, Universiteit Gent \\
rcardon@inf.uc3m.es, as.dogruoz@ugent.be\\
}

\abstract{ Readability is a key concept in the current era of abundant written information. To help making texts more readable and make information more accessible to everyone, a line of researched aims at making texts accessible for their
target audience: automatic text simplification (ATS). Lately, there have been studies on the correlations between
automatic evaluation metrics in ATS and human judgment. However, the correlations between
those two aspects and commonly available readability measures (such as readability formulas or linguistic features) have not been the
focus of as much attention. In this work, we investigate the place of
readability measures in ATS by complementing the existing studies on evaluation
metrics and human judgment, on English. We first discuss the relationship between ATS and research
in readability, then we report a study on correlations between readability
measures and human judgment, and between readability measures and ATS evaluation
metrics. We identify that in general, readability measures do not correlate well with automatic metrics and human judgment. We argue that as the three different angles from which simplification can be assessed tend to exhibit rather low correlations with one another, there is a need for a clear definition of the construct in ATS. \\
\newline
 }

\begin{document}
  \maketitleabstract

  \section{Introduction}
  The accessibility of written information is an important question: outside natural
  language processing (NLP), domains like medicine \cite{gu2024systematic} or business
  \cite{huong2024annual} have been studying the readability of the documents
  they produce (e.g. medical reports or information for patients, business reports
  for shareholders). Usually, those studies are performed using traditional
  readability formulas, like the Flesch Reading Ease \cite{flesch1948new}
  or Dale-Chall \cite{dale1948formula} formulas. Recently, they have been acknowledging
  the reliability issues that come with those formulas \cite{alzaid2024limitations}.
  In NLP, \label{sec:intro} Automatic text simplification
  (ATS) is a task that aims at transforming texts
  in order to make them more accessible, while preserving their meaning
  \cite{saggion2017automatic}. In ATS studies, the goal is sometimes phrased as
  increasing the readability of a text. However, there is a knowledge gap as to how readability measures and judgment on ATS systems (either automated or human) correlate. In this paper, we investigate the place
  that readability occupies in the ATS landscape. We analyze the discourse on readability
  in ATS works by putting it in contrast with automatic readability
  assessment (ARA), that aims at identifying the readability level of texts
  \cite{vajjala-2022-trends}. While readability is regularly mentioned in current  ATS works, ATS does not leverage ARA developments.

  We also study the correlations between readability measures, human
  judgment and ATS evaluation metrics. While there have been studies about the
  correlations between ATS evaluation metrics and human judgment
  \cite{alva-manchego-etal-2021-un, cripwell2024evaluating}, the correlations between
  those two aspects and commonly available readability measures have not been the
  focus of as much attention. We fill this gap by reporting a study on
  correlations between readability measures and human judgment, and between readability
  measures and ATS evaluation metrics.

  Our contributions are the following: a discussion of ATS and ARA that identifies
  the bridges that remain to be made between the two fields; experiments with
  readability measures for ATS evaluation that fill a knowledge gap regarding
  correlations of evaluation practices and human judgment; insights for future developments of the field of ATS.

  \section{Related Work}
  \label{sec:sota}

  In this section, we discuss the fields of readability and text simplification
  that we introduce separately (Sections \ref{sec:read} and \ref{sec:ts}) before
  discussing how the two fields interact with each other (Section \ref{sec:read-ts}).

  \subsection{Readability}
  \label{sec:read}

  Readability is a field of research that is considered to date back to the 1920's,
  with the first attempt to quantify the readability of English texts \citet{lively1923method}.
  This first method relied on a list of word frequencies
  \cite{thorndike1921word}, relying on the assumption that texts made of frequent words are more readable. \citet{franccois2015readability}
  distinguishes several eras in text readability research, from \citet{lively1923method}
  to various paradigms of ``AI readability". We give an overview of this historical
  perspective below.

  The early period consisted of identifying predictors and tuning coefficient weights out of corpus-based observations and annotations from a given target
  audience. The most famous examples for English are Flesch Reading Ease \cite[FRE]{flesch1948new}
  and Flesch-Kincaid Grade Level \cite[FKGL]{kincaid1975derivation}, which rely
  on word count and number of syllables per word.

  The first approaches to measuring readability with NLP tools relied on linear
  regression on linguistic (i.e. syntactic and lexical) variables \cite{daoust1996sato},
  latent semantic analysis for textual coherence and cohesion
  \cite{foltz1998measurement} and probabilities computed with language modeling
  \cite{si2001statistical}.

  \citet{franccois2015readability} concludes by noting an emerging trend at the time
  in ARA, that relies on automatic feature extraction using neural
  networks. Ten years later, ARA has developed into a lively line of research
  \cite{vajjala-2022-trends}. ARA has been explored with distributional text representations
  and with linguistic features. The distributional text representations follow
  the advancements of research in machine learning, notably with the development
  of transformers \cite{vaswani2017attention}. Regarding linguistic features, the
  way to select and leverage them is still an open question. Nonetheless, research
  on this question is facilitated by the appearance of tools that can be used to compute
  an increasingly high number of features, for example for English
  \cite{kyle2021assessing, kyle2018tool, lu2010automatic, crossley2019tool} or
  French \cite{wilkens-etal-2022-fabra}. These tools produce raw analyses of texts with hundreds
  of features, with no recommendations as to how to select and use them which is
  left up to their user. This knowledge gap has fueled research, notably with works that aim at
  combining those numeric representations with distributional representations
  \cite{deutsch-etal-2020-linguistic, lee-etal-2021-pushing, wilkens-etal-2024-exploring}.

  The readability features depend heavily on the language that is under study. The aforementioned tools rely on language-dependent resource such as reference
  corpora, vocabulary lists, or pre-trained models (e.g. for POS-tagging or
  syntactic analysis).

  \subsection{Automatic Text Simplification}
  \label{sec:ts}

  In this section, we briefly describe ATS to lay the ground for the discussion of
  how it integrates considerations about readability that comes in the next section
  (Section \ref{sec:read-ts}).

  \paragraph{Methods}
  ATS has traditionally been performed at the sentence-level
  \cite{saggion2017automatic}. In the early works, the goal was to make sentences simpler to
  handle as an input for other NLP systems such as syntactic parsers \cite{chandrasekar-etal-1996-motivations}.
  It was only later explored as a means of simplifying texts to make them easier
  to understand by humans \cite{carroll-etal-1999-simplifying}. These initial methods
  were rule-based and targeted specific operations \cite{cardon-bibal-2023-operations}
  (e.g., removing appositive clauses, changing the voice of a sentence from passive
  to active). The recent developments of generative models has accelerated the
  shift of ATS research to document-level simplification
  \cite{sun-etal-2021-document}, notably with multi-agent architectures \cite{mo-hu-2024-expertease,
  fang-etal-2025-collaborative} while sentence simplification is still being explored
  \cite{kew-etal-2023-bless}.

  \paragraph{Evaluation.}
  Evaluation of ATS is an open question. Traditional readability, mostly FKGL or
  adaptions of FRE for other languages are often reported, while it has been shown
  that they correlate poorly with human judgment on simplification \cite{tanprasert-kauchak-2021-flesch,alva-manchego-etal-2021-un}.
  For sentence simplification, the most common metrics are BLEU \cite{papineni-etal-2002-bleu},
  SARI \cite{xu-etal-2016-optimizing}, with an adaptation for document-level simplification
  D-SARI \cite{sun-etal-2021-document}, and BERTScore
  \cite{zhang2020bertscoreevaluatingtextgeneration}. BLEU and BERTScore compare the
  output to one or more references, while (D-)SARI adds the input into the
  computation. Their correlation with the task is also unclear \cite{alva-manchego-etal-2021-un,sulem-etal-2018-bleu},
  although BLEU is often interpreted as an indicator of meaning preservation, SARI
  of simplicity, and BERTScore of meaning preservation and fluency.

  These three indicators are the three criteria that are also used for human judgment
  to evaluate sentence simplification, typically on 5-point Likert scales. For
  document-level simplification, human evaluation is not stabilized. \citet{cripwell2024evaluating}
  use the same criteria with binary questions instead of Likert scales.
  \citet{sun-etal-2021-document} ask judges to evaluate ``overall simplicity"
  that they define as simplicity with other quality criteria such as ease of
  reading and meaning preservation. \citet{vasquez-rodriguez-etal-2023-document}
  ask judges to evaluate textual coherence. \citet{agrawal2024text} evaluate
  meaning preservation by studying human performance on reading comprehension
  tests.

  \subsection{Readability and Text Simplification}
  \label{sec:read-ts}

  In most works where readability and ATS interact, readability is leveraged through linguistic
  features to give information about datasets \cite{battisti-etal-2020-corpus, vajjala-lucic-2018-onestopenglish,
  yaneva-etal-2016-evaluating, stajner-saggion-2013-readability,dellorletta-etal-2011-read,
  aluisio-etal-2010-readability}. Other research works \citet{jingshen-etal-2024-readability}
  rely on features for data selection instead. Readability features, in conjunction
  with similarity measures, are leveraged to mine sentence pairs to produce a
  parallel corpus for Chinese idiom simplification. \citet{de-martino-2023-processing}
  investigates the link between eye-tracking data and readability features on
  Italian data. While it is a preliminary study, it suggests that eye-tracking
  is a promising method for evaluating the effect of simplification transformations.

  Some ATS studies use readability features or metrics in their evaluation
  protocol. \citet{scholz-wenzel-2025-evaluating} evaluate 18 readability
  features (i.e., syntactic, POS-based, semantic and fluency features) for English and
  German text simplification. Their findings are that some measures are transferable (semantic and fluency features), and that the behavior of statistical, POS-based and
  syntactic metrics seem to be strongly language-dependent. \citet{paula-camilo-junior-2024-evaluating}
  use a Portuguese adaption of FRE as an evaluation metric for ATS. \cite{engelmann-etal-2024-arts}
  use the FRE and Dale-Chall formulas to perform pairwise comparisons in an Elo-like
  ranking system. They compare it to human judgments and GPT 3.5 performance. They
  find that Dale-Chall has the highest correlation to human judgment, above GPT
  3.5, while FRE obtains the lowest correlations.

  Readability can also be incorporated in ATS methods. \citet{flores-etal-2023-medical}
  use a bounded FKGL (ranging from 4 to 20, based on empirical observations) as
  a component of their loss in a neural model for text simplification. \cite{maddela-alva-manchego-2025-adapting}
  prompt LLMs for document-level simplification by including CEFR levels in the prompt,
  as was also done by \citet{imperial-tayyar-madabushi-2023-flesch}. Using CEFR
  as a proxy for readability was initiated with the release of the CEFR-SP
  dataset \cite{arase-etal-2022-cefr}, a corpus of 17,000 English sentences annotated
  with CEFR levels.

  Lexical complexity features have been leveraged for lexical simplification \cite{north2025deep}.
  \citet{hazim-etal-2022-arabic} introduce a system that highlights complex words
  in a text editor to help humans manually simplify texts. \citet{maddela-xu-2018-word}
  use lexical features to rank candidates for substitution in a neural lexical simplification
  system. \cite{grigonyte-etal-2014-improving} rely on features to perform
  complex word identification.

  In conclusion, we observe that different approaches to readability (features, formulas,
  eye-tracking, CEFR levels) are explored in ATS works. The two approaches that are
  widely present in ATS are traditional formulas, which have consistently been
  used as an evaluation metric, and readability features, that have been used to
  give information about datasets. In this study, we explore how features
  correlate with human judgment on the simplification task, for the English language.

  \section{Readability Measures and ATS Metrics}

  \subsection{Data}

  In order to study how readability features correlate with the evaluation
  protocols in ATS, we rely on English data that are labeled with human judgment and on which
  automatic metrics can be computed. Two studies provide this kind of data, at the
  sentence level \cite{alva-manchego-etal-2021-un} and at the document level \cite{maddela-alva-manchego-2025-adapting}.
  Both studies aim at studying the link between automatic metrics and human
  judgment. In this paper, we add observations on the link between readability
  measures and human judgment, and on the link between readability measures and automatic  metrics. We describe the datasets below.

  \textbf{SimplicityDA.} For the sentence-level study, we use Simplicity-DA
  \cite{alva-manchego-etal-2021-un}\footnote{\url{https://github.com/feralvam/metaeval-simplification}}.
  It is a set of 600 sentence simplification system outputs in English, each one
  annotated by 15 crowdworkers along with the three common human judgment criteria ( fluency, simplicity and meaning preservation) in
  ATS on a 0-100 scale, using the
  direct assessment method. The dataset also includes automatic scores for each
  sentence: BLEU, SARI, BERTScore and SAMSA.

  \textbf{D-Wikipedia.} For the document-level study, we use D-Wikipedia \cite{sun-etal-2021-document}.
  D-Wikipedia is a corpus of aligned paragraph pairs that come from the English
  Wikipedia for the complex side and Simple English Wikipedia for the simple
  side. \citet{maddela-alva-manchego-2025-adapting} released a subset of 100
  paragraph pairs from D-Wikipedia, each with 4 simplifications produced by automatic
  systems, resulting in 500 paragraph pairs. Those 500 pairs were rated by three
  human judges on fluency, simplicity and meaning preservation on a 5-point
  Likert scale. We compute the automatic metrics values with the code provided with
  the dataset\footnote{\url{https://github.com/cardiffnlp/document-simplification}}.
  Those automatic metrics are BLEU, SARI, D-SARI, BERTScore and LENS. \citet{maddela-alva-manchego-2025-adapting}
  also introduce adaptations of SARI, LENS and BERTScore (respectively Agg-SARI,
  Agg-LENS and Agg-BERTScore) to the document-level simplification task by
  aggregating scores computed at the sentence-level.

  \subsection{Readability Measures}

  \label{sec:material}

  \begin{figure*}[!ht]
    \centering
    \begin{subfigure}
      [b]{0.45\textwidth}
      \includegraphics[width=\textwidth, trim=0 150 0 150, clip]{
        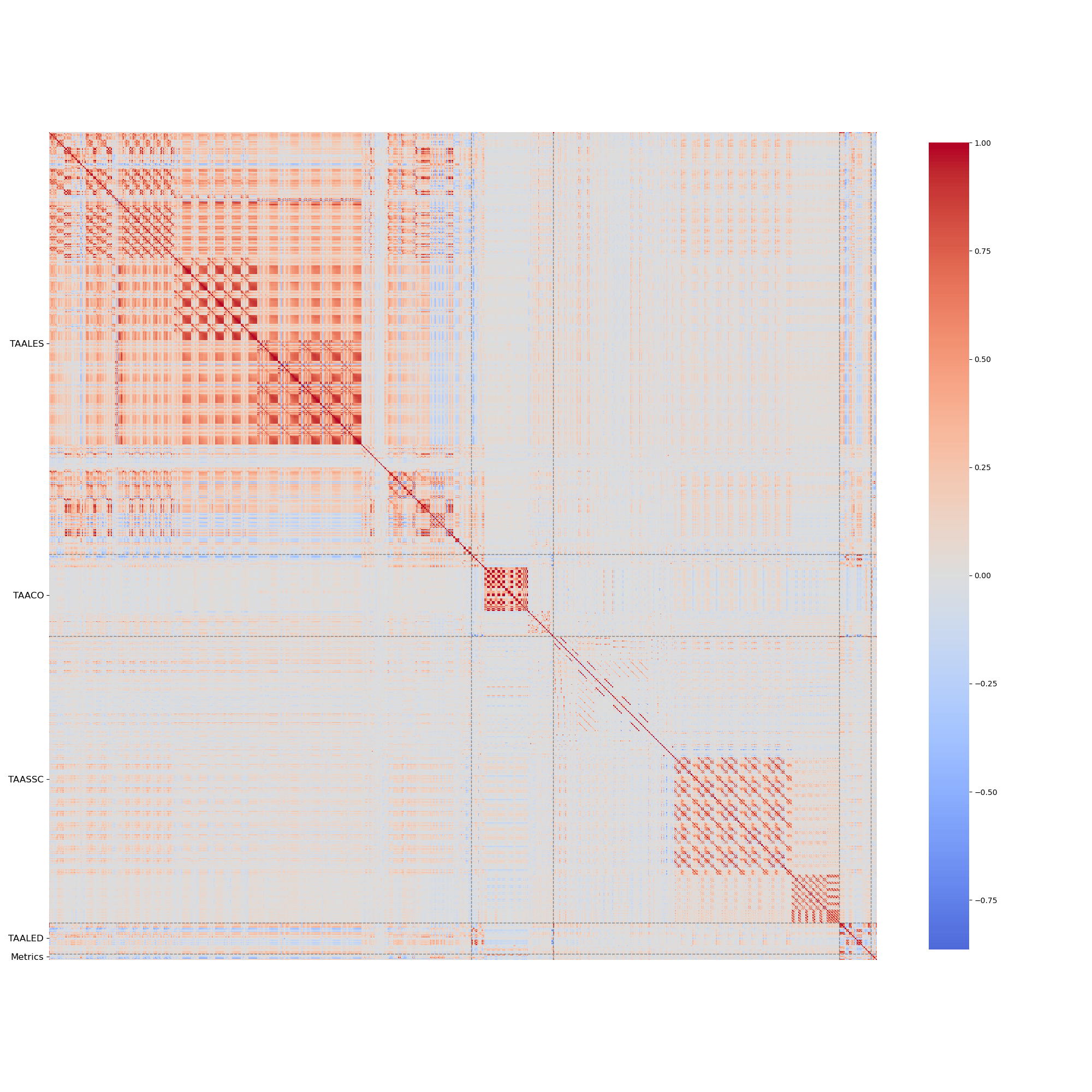
      }
      \caption{Difference between original and simple, \\
      SimplicityDA (sentence-level).}
      \label{fig:heat_sent_diff}
    \end{subfigure}
    \begin{subfigure}
      [b]{0.45\textwidth}
      \includegraphics[width=\textwidth, trim=0 150 0 150, clip]{
        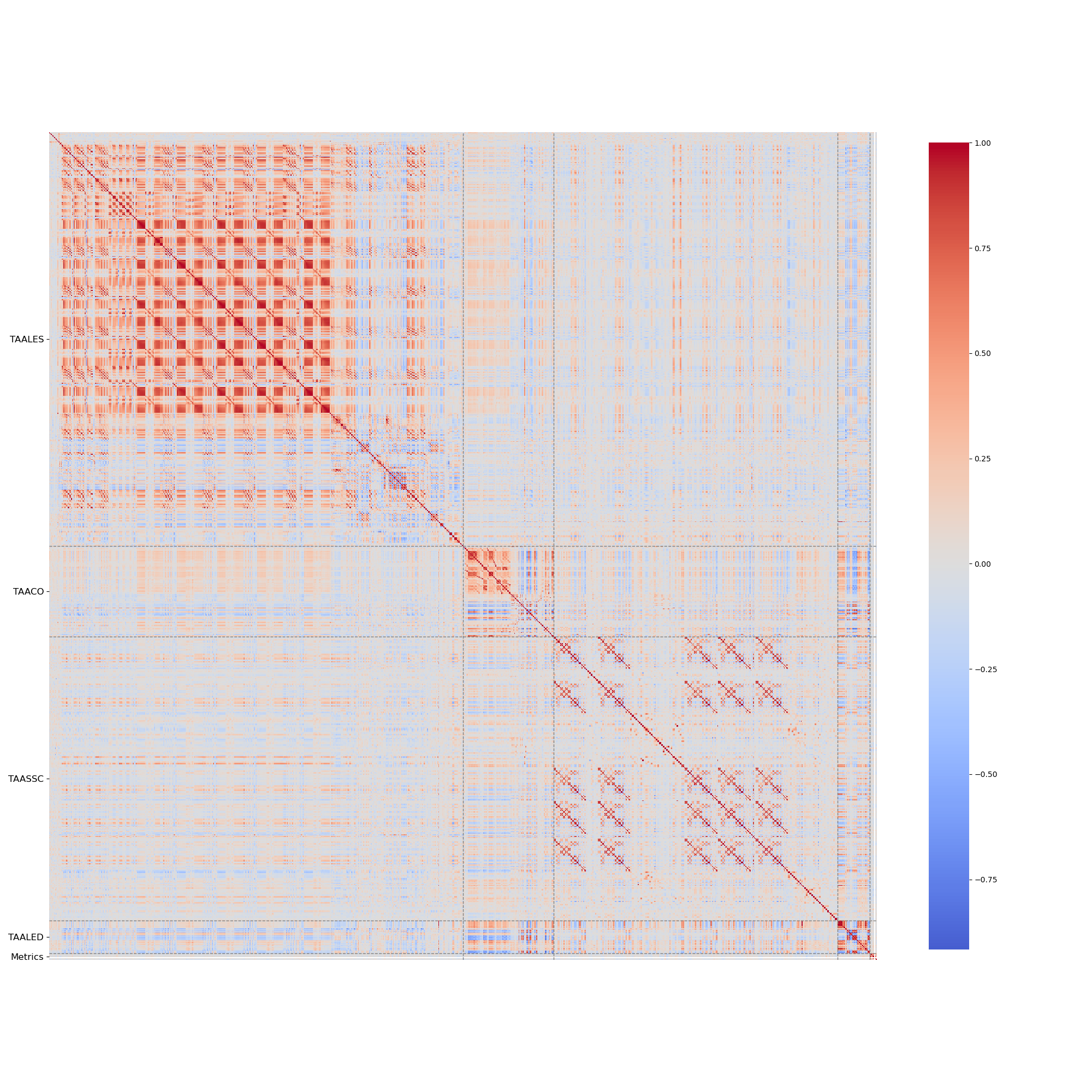
      }
      \caption{Difference between original and simple, \\
      D-Wikipedia (document-level).}
      \label{fig:heat_doc_diff}
    \end{subfigure}
    \\
    \begin{subfigure}
      [b]{0.45\textwidth}
      \includegraphics[width=\textwidth, trim=0 150 0 150, clip]{
        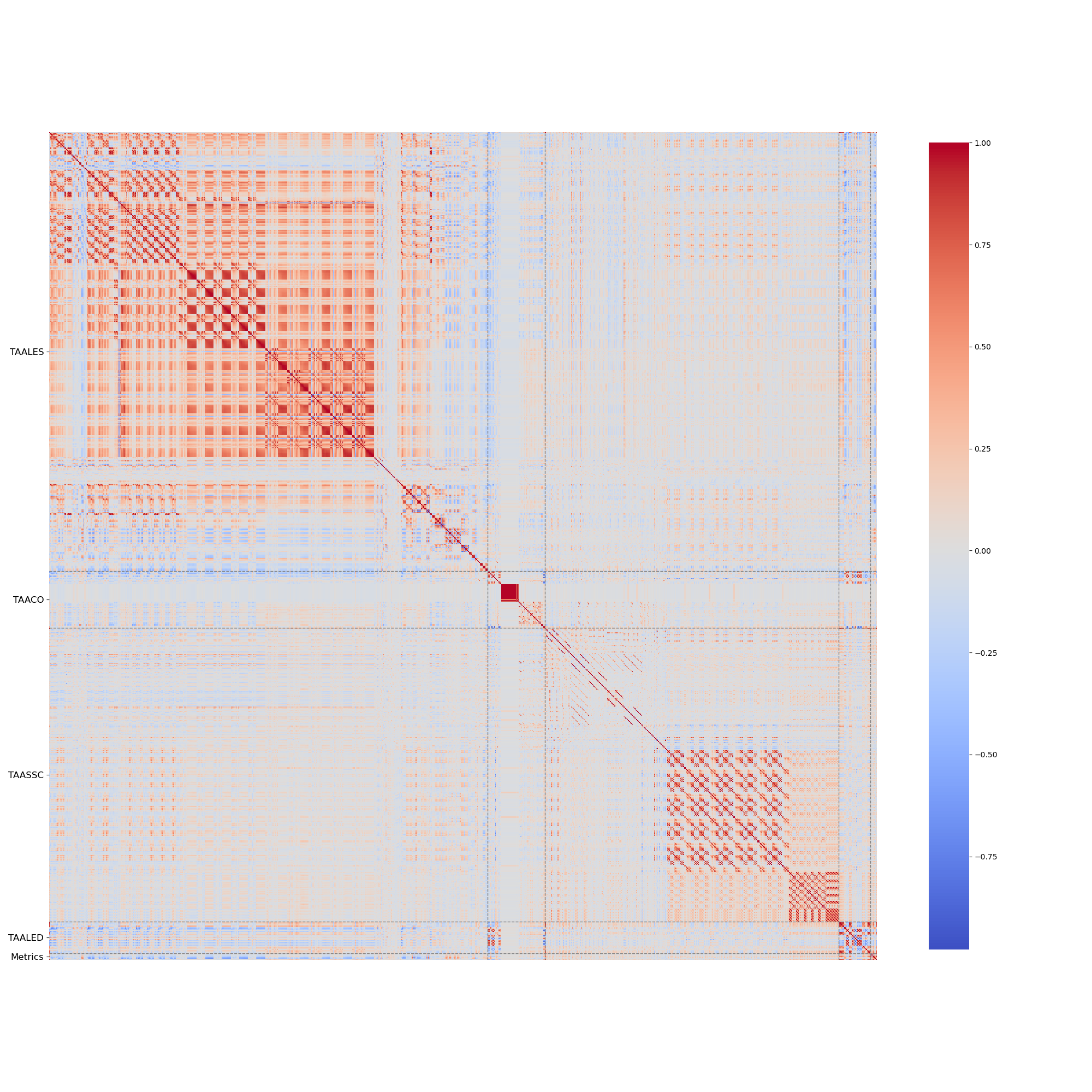
      }
      \caption{Simple side of SimplicityDA (sentence-level).}
      \label{fig:heat_sent_simp}
    \end{subfigure}
    \begin{subfigure}
      [b]{0.45\textwidth}
      \includegraphics[width=\textwidth, trim=0 150 0 150, clip]{
        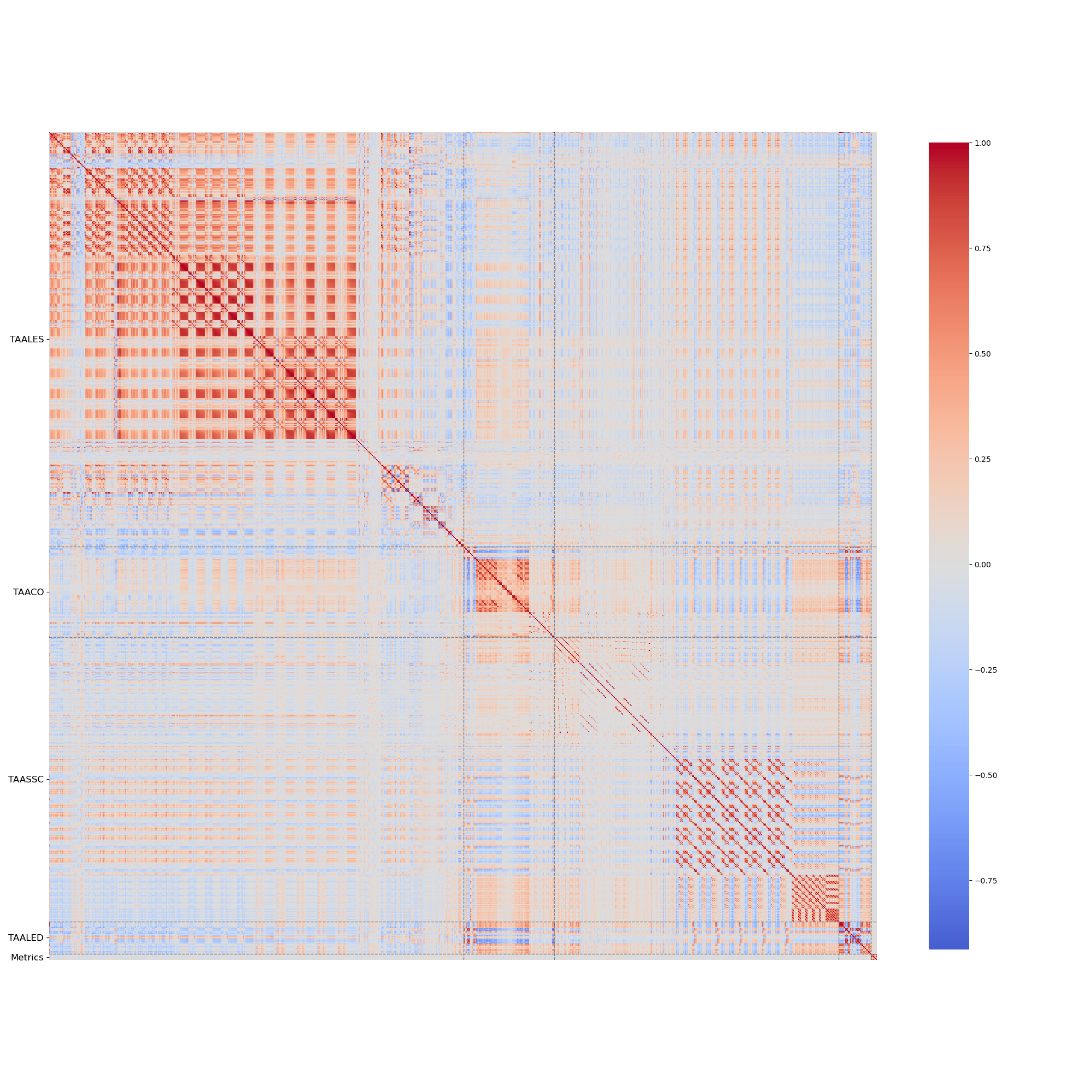
      }
      \caption{Simple side of D-Wikipedia (document-level).}
      \label{fig:heat_doc_simp}
    \end{subfigure}
    \caption{Pearson correlation matrices of readability measures and metrics. Dashed
    lines indicate the boundaries of feature groups (from top to bottom, and the
    same from left to right: TAALES, TAACO, TAASSC, TAALED, and Metrics).}
    \label{fig:matrices}
  \end{figure*}

  \paragraph{Readability Features.}
  As discussed in Section \ref{sec:sota}, readability is now mostly explored
  with two types of text representations: distributional embeddings and textual features.
  As distributional embeddings are leveraged for ATS methods and evaluation, we focus
  on textual features. To compute these features, we use the four tools (see below to implement a total of 1,066 readability-related features for
  English:
  \begin{itemize}
    \item TAALED \cite{kyle2021assessing}\footnote{\url{https://www.linguisticanalysistools.org/taaled.html}}
      computes 38 features related to lexical diversity, such as different type-token ratios or MTLD (Measures of Textual Lexical Diversity).

    \item TAALES \cite{kyle2018tool}\footnote{\url{https://www.linguisticanalysistools.org/taales.html}}
      computes 484 features related to lexical sophistication. Many of those features
      are variations around word frequency, computed on various corpora such as
      BNC \cite[The British National Corpus]{british-national-corpus} and COCA
      \cite[The Corpus of Contemporary American English]{COCA} for example. Other
      features are related to neighborhood (e.g. orthographic, phonological),
      age of acquisition, psycholinguistic norms (e.g. concreteness,
      imageability, meaningfulness).

    \item TAASSC \cite{lu2010automatic}\footnote{\url{https://www.linguisticanalysistools.org/taassc.html}}
      computes 376 features related to syntactic sophistication. Those features rely
      on dependency analysis and part-of-speech tagging. Some examples of
      features are conjunctions per clause, verbal modifiers per nominal,
      frequency of constructions compared to references coming from different
      corpora (e.g. BNC, COCA and others) or more traditional ones such as
      average sentence length.

    \item TAACO \cite{crossley2019tool}\footnote{\url{https://www.linguisticanalysistools.org/taaco.html}}
      computes 168 features related to cohesion. Some examples are semantic similarity
      between word2vec \cite{mikolov2013efficient} embeddings of adjacent sentences,
      or token overlap between adjacent sentences or paragraphs.
  \end{itemize}
  The complete list of features and their formulas is available in the documentation
  of each tool.

  \paragraph{Readability Metrics.}
  We also compute the following set of traditional readability metrics for
  English, using the \href{https://pypi.org/project/textstat/}{\texttt{textstat}}
  Python library: Flesch Reading Ease \cite{flesch1948new}, Dale-Chall \cite{dale1948formula},
  Gunning-Fog \cite{gunning1952technique}, Linsear Write \cite{o1966gobbledygook},
  ARI \cite{smith1967automated}, SMOG \cite{mc1969smog}, Flesch-Kincaid Grade
  Level \cite{kincaid1975derivation}, and Coleman-Liau \cite{coleman1975computer}.

  \begin{table*}
    [h!t]
    \begin{small}
      \centering
      \begin{subtable}
        [t]{0.32\textwidth}
        \centering
        \caption{simp -- simplicity}
        \begin{tabular}{lr}
          \toprule Variable          & $r$                         \\
          \midrule log\_ttr\_aw & 0.199  \\
          log\_ttr\_cw          & 0.193  \\
          McD\_CD\_FW           & 0.178  \\
          basic\_connectives    & -0.173 \\
          lemma\_ttr            & 0.173  \\
          lemma\_mattr          & 0.173  \\
          MRC\_Familiarity\_CW  & 0.168  \\
          msttr50\_aw           & 0.164  \\
          mattr50\_aw           & 0.164  \\
          bigram\_lemma\_ttr    & 0.159  \\
          \bottomrule
        \end{tabular}
      \end{subtable}\hfill
      \begin{subtable}
        [t]{0.32\textwidth}
        \centering
        \caption{simp -- fluency}
        \begin{tabular}{lr}
          \toprule Variable                             & $r$                         \\
          \midrule \texttt{COCA\_magazine\_bi\_MI} & 0.262  \\
          {log\_ttr\_aw}                           & 0.242  \\
          {COCA\_news\_bi\_MI}                     & 0.237  \\
          {COCA\_fiction\_bi\_MI}                  & 0.237  \\
          {basic\_connectives}                     & -0.218 \\
          {COCA\_spoken\_bi\_MI}                   & 0.217  \\
          {log\_ttr\_cw}                           & 0.217  \\
          {conjunctions}                           & -0.203 \\
          {conj\_per\_cl}                          & -0.195 \\
          {acad\_lemma\_attested}                  & 0.195  \\
          \bottomrule
        \end{tabular}
      \end{subtable}\hfill
      \begin{subtable}
        [t]{0.32\textwidth}
        \centering
        \caption{simp -- meaning}
        \begin{tabular}{lr}
          \toprule Variable                    & $r$                        \\
          \midrule \texttt{root\_ttr\_cw} & 0.295 \\
          {root\_ttr\_aw}                 & 0.286 \\
          {log\_ttr\_cw}                  & 0.269 \\
          {basic\_ncontent\_types}        & 0.254 \\
          {hyper\_verb\_noun\_Sav\_P1}    & 0.239 \\
          {mtld\_ma\_wrap\_aw}            & 0.234 \\
          {hyper\_verb\_noun\_Sav\_Pav}   & 0.233 \\
          {hyper\_verb\_noun\_s1\_p1}     & 0.232 \\
          {linsear}                       & 0.231 \\
          {basic\_ntypes}                 & 0.231 \\
          \bottomrule
        \end{tabular}
      \end{subtable}
      \bigskip
      \begin{subtable}
        [t]{0.32\textwidth}
        \centering
        \caption{delta -- simplicity}
        \begin{tabular}{lr}
          \toprule Variable                   & $r$                         \\
          \midrule \texttt{conjunctions} & 0.215  \\
          {basic\_connectives}           & 0.208  \\
          {av\_pobj\_deps\_NN}           & -0.178 \\
          {log\_ttr\_cw}                 & -0.176 \\
          {adv\_ttr}                     & -0.169 \\
          {log\_ttr\_aw}                 & -0.167 \\
          {av\_pobj\_deps}               & -0.159 \\
          {MRC\_Familiarity\_CW}         & -0.153 \\
          {MRC\_Imageability\_CW}        & -0.152 \\
          {hyper\_verb\_noun\_Sav\_P1}   & -0.150 \\
          \bottomrule
        \end{tabular}
      \end{subtable}\hfill
      \begin{subtable}
        [t]{0.32\textwidth}
        \centering
        \caption{delta -- fluency}
        \begin{tabular}{lr}
          \toprule Variable                    & $r$                         \\
          \midrule \texttt{root\_ttr\_aw} & -0.300 \\
          {root\_ttr\_cw}                 & -0.283 \\
          {basic\_ntypes}                 & -0.242 \\
          {conjunctions}                  & 0.239  \\
          {mtld\_ma\_wrap\_aw}            & -0.237 \\
          {basic\_ncontent\_types}        & -0.235 \\
          {basic\_connectives}            & 0.226  \\
          {av\_pobj\_deps\_NN}            & -0.212 \\
          {log\_ttr\_aw}                  & -0.212 \\
          {root\_ttr\_fw}                 & -0.211 \\
          \bottomrule
        \end{tabular}
      \end{subtable}\hfill
      \begin{subtable}
        [t]{0.32\textwidth}
        \centering
        \caption{delta -- meaning}
        \begin{tabular}{lr}
          \toprule Variable                    & $r$                         \\
          \midrule \texttt{root\_ttr\_aw} & -0.426 \\
          {root\_ttr\_cw}                 & -0.398 \\
          {basic\_ntypes}                 & -0.392 \\
          {nwords}                        & -0.390 \\
          {Word Count}                    & -0.383 \\
          {basic\_ntokens}                & -0.373 \\
          {basic\_ncontent\_tokens}       & -0.365 \\
          {mtld\_ma\_wrap\_aw}            & -0.363 \\
          {basic\_ncontent\_types}        & -0.361 \\
          {basic\_nfunction\_types}       & -0.350 \\
          \bottomrule
        \end{tabular}
      \end{subtable}
    \end{small}
    \caption{Top absolute values of significant correlation coefficients (\textit{p < .001}) between human judgment on the simplicityDA dataset and readability measures.}
    \label{tab:simfeatures}
  \end{table*}

\begin{table*}
  [h!]
  \begin{small}
    \centering
    \resizebox{\textwidth}{!}{{}
      \begin{subtable}[t]{0.32\textwidth}
        \centering
        \caption{simp -- simplicity}
        \begin{tabular}{lr}
          \toprule Variable & $r$ \\
          \midrule
          \scriptsize
          Word Count & \scriptsize -0.205  \\
          \scriptsize {Kuperman\_AoA\_FW} & \scriptsize -0.138  \\
          \scriptsize {Phono\_N\_FW} & \scriptsize 0.116  \\
          \scriptsize {Phono\_N\_H\_FW} & \scriptsize 0.113  \\
          \scriptsize {hyper\_noun\_S1\_P1} & \scriptsize 0.110  \\
          \scriptsize {hyper\_noun\_Sav\_Pav} & \scriptsize 0.102  \\
          \scriptsize {MRC\_Familiarity\_FW} & \scriptsize 0.096  \\
          \scriptsize {COCA\_fiction\_Range\_CW} & \scriptsize -0.095  \\
          \scriptsize {BNC\_Spoken\_3gram\_NF} & \scriptsize -0.093  \\
          \scriptsize {poly\_adj} & \scriptsize -0.091  \\
          \bottomrule
        \end{tabular}
      \end{subtable}
      \hfill
      \begin{subtable}[t]{0.32\textwidth}
        \centering
        \caption{simp -- grammar}
        \begin{tabular}{lr}
          \toprule Variable & $r$ \\
          \midrule
          \scriptsize {COCA\_spoken\_Trigram\_Frequency\_Log} & \scriptsize 0.103  \\
          \scriptsize {WN\_SD\_CW} & \scriptsize -0.101  \\
           \scriptsize {COCA\_news\_tri\_2\_DP} & \scriptsize 0.099  \\
          \scriptsize {Brysbaert\_CC\_AW} & \scriptsize 0.098  \\
          \scriptsize {COCA\_magazine\_tri\_2\_MI2} & \scriptsize 0.098  \\
          \scriptsize {COCA\_magazine\_tri\_2\_DP} & \scriptsize 0.098  \\
          \scriptsize {COCA\_spoken\_tri\_prop\_20k} & \scriptsize 0.098  \\
          \scriptsize {OG\_N\_H} & \scriptsize -0.098  \\
          \scriptsize {Freq\_N\_OGH} & \scriptsize -0.098  \\
          \scriptsize {COCA\_news\_tri\_prop\_10k} & \scriptsize 0.096  \\
          \bottomrule
        \end{tabular}
      \end{subtable}
      \hfill
      \begin{subtable}[t]{0.32\textwidth}
        \centering
        \caption{simp -- meaning}
        \begin{tabular}{lr}
          \toprule Variable & $r$ \\
          \midrule
          \scriptsize Word Count & \scriptsize -0.433  \\
          \scriptsize {Kuperman\_AoA\_FW} & \scriptsize -0.243  \\
          \scriptsize {Brown\_Freq\_CW} & \scriptsize 0.240  \\
          \scriptsize {TL\_Freq\_CW} & \scriptsize 0.239  \\
          \scriptsize {KF\_Freq\_CW} & \scriptsize 0.219  \\
          \scriptsize {OLDF\_FW} & \scriptsize 0.213  \\
          \scriptsize {Freq\_N\_OG\_CW} & \scriptsize 0.211  \\
          \scriptsize {OG\_N\_H\_CW} & \scriptsize 0.200  \\
          \scriptsize {Freq\_N\_OGH\_CW} & \scriptsize 0.200  \\
          \scriptsize {poly\_adj} & \scriptsize -0.192  \\
          \bottomrule
        \end{tabular}
      \end{subtable}
    }
    \bigskip
    \resizebox{\textwidth}{!}{
      \begin{subtable}[t]{0.32\textwidth}
        \centering
        \caption{delta -- simplicity}
        \begin{tabular}{lr}
          \toprule Variable & $r$ \\
          \midrule
          \scriptsize
          \scriptsize Word Count & \scriptsize 0.298  \\
          \scriptsize {TL\_Freq\_FW\_Log} & \scriptsize -0.267  \\
          \scriptsize {COCA\_fiction\_Frequency\_Log\_FW} & \scriptsize -0.258  \\
          \scriptsize {KF\_Freq\_FW\_Log} & \scriptsize -0.252  \\
          \scriptsize {BNC\_Written\_Freq\_FW\_Log} & \scriptsize -0.249  \\
          \scriptsize {COCA\_news\_Frequency\_Log\_FW} & \scriptsize -0.245  \\
          \scriptsize {COCA\_magazine\_Frequency\_Log\_FW} & \scriptsize -0.239  \\
          \scriptsize {AWL\_Sublist\_5\_Normed} & \scriptsize 0.237  \\
          \scriptsize {BNC\_Spoken\_Freq\_FW\_Log} & \scriptsize -0.236  \\
          \scriptsize {Brown\_Freq\_FW\_Log} & \scriptsize -0.234  \\
          \bottomrule
        \end{tabular}
      \end{subtable}
      \hfill
      \begin{subtable}[t]{0.32\textwidth}
        \centering
        \caption{delta -- grammar}
        \begin{tabular}{lr}
          \toprule Variable & $r$ \\
          \midrule
          \scriptsize
          \scriptsize {AWL\_Sublist\_10\_Normed} & \scriptsize -0.291  \\
          \scriptsize {COCA\_fiction\_Frequency\_FW} & \scriptsize -0.216  \\
          \scriptsize {BNC\_Spoken\_3gram\_NF\_Log} & \scriptsize 0.214  \\
          \scriptsize {TL\_Freq\_FW} & \scriptsize -0.202  \\
          & \\
          & \\
          & \\
          & \\
          & \\
          & \\
          \bottomrule
        \end{tabular}
      \end{subtable}
      \hfill
      \begin{subtable}[t]{0.32\textwidth}
        \centering
        \caption{delta -- meaning}
        \begin{tabular}{lr}
          \toprule Variable & $r$ \\
          \midrule
          \scriptsize
          \scriptsize Word Count & \scriptsize 0.342  \\
          \scriptsize {Kuperman\_AoA\_AW} & \scriptsize 0.270  \\
          \scriptsize {COCA\_spoken\_Frequency\_Log\_CW} & \scriptsize -0.255  \\
          \scriptsize {PLDF\_FW} & \scriptsize -0.248  \\
          \scriptsize {COCA\_spoken\_RL\_CW} & \scriptsize -0.228  \\
          \scriptsize {SUBTLEXus\_Range\_FW} & \scriptsize -0.225  \\
          \scriptsize {COCA\_news\_RL\_FW} & \scriptsize -0.224  \\
          \scriptsize {COCA\_spoken\_RL\_FW} & \scriptsize -0.221  \\
          \scriptsize {KF\_Ncats\_FW} & \scriptsize -0.220  \\
          \scriptsize {Kuperman\_AoA\_CW} & \scriptsize 0.211  \\
          \bottomrule
        \end{tabular}
      \end{subtable}
    }
  \end{small}
  \caption{Top absolute values of significant correlation coefficients (\textit{p < .05}) between human judgment on the DWiki dataset and features.}
  \label{tab:wikifeatures}
\end{table*}

     \normalsize

  \begin{table*}
    [ht!]
          \begin{footnotesize}
    \begin{center}
      \begin{footnotesize}
        \begin{subtable}
          [t]{0.237\textwidth}
          \centering
          \caption{simp -- bertscore\_F1}
          \begin{tabular}{lr}
            \toprule Variable                & $r$                        \\
            \midrule 
            \scriptsize root\_ttr\_aw      & \scriptsize 0.339 \\
            \scriptsize rootTTRCW               & \scriptsize 0.317 \\
            \scriptsize log\_ttr\_cw                & \scriptsize 0.311 \\
            \scriptsize mtld\_ma\_wrap\_aw          & \scriptsize 0.294 \\
            \scriptsize hyper\_verb\_noun\_Sav\_P1  & \scriptsize 0.291 \\
            \scriptsize hyper\_verb\_noun\_Sav\_Pav & \scriptsize 0.283 \\
            \scriptsize hyper\_verb\_noun\_s1\_p1   & \scriptsize 0.279 \\
            \scriptsize log\_ttr\_aw                & \scriptsize 0.277 \\
            \scriptsize hyper\_noun\_S1\_P1         & \scriptsize 0.276 \\
            \scriptsize basic\_ntypes               & \scriptsize 0.273 \\
            \bottomrule
          \end{tabular}
        \end{subtable}\hfill
        \begin{subtable}
          [t]{0.237\textwidth}
          \centering
          \caption{simp -- bleu}
          \begin{tabular}{lr}
            \toprule Variable           & $r$                        \\
            \midrule 
            \scriptsize root\_ttr\_aw & \scriptsize 0.352 \\
            \scriptsize root\_ttr\_cw          & \scriptsize 0.341 \\
            \scriptsize linsear                & \scriptsize 0.329 \\
            \scriptsize mtld\_ma\_wrap\_aw     & \scriptsize 0.324 \\
            \scriptsize basic\_ntypes          & \scriptsize 0.315 \\
            \scriptsize Word Count             & \scriptsize 0.314 \\
            \scriptsize nwords                 & \scriptsize 0.313 \\
            \scriptsize basic\_ncontent\_types & \scriptsize 0.303 \\
            \scriptsize log\_ttr\_cw           & \scriptsize 0.295 \\
            \scriptsize fkgl                   & \scriptsize 0.290 \\
            \bottomrule
          \end{tabular}
        \end{subtable}\hfill
        \begin{subtable}
          [t]{0.237\textwidth}
          \centering
          \caption{simp -- samsa}
          \begin{tabular}{lr}
            \toprule Variable                 & $r$                         \\
            \midrule 
            \scriptsize cl\_ndeps\_std\_dev & \scriptsize -0.309 \\
            \scriptsize basic\_ntokens               & \scriptsize -0.302 \\
            \scriptsize Word Count                   & \scriptsize -0.285 \\
            \scriptsize basic\_ntypes                & \scriptsize -0.284 \\
            \scriptsize basic\_nfunction\_tokens     & \scriptsize -0.277 \\
            \scriptsize nwords                       & \scriptsize -0.277 \\
            \scriptsize basic\_ncontent\_tokens      & \scriptsize -0.271 \\
            \scriptsize mtld\_ma\_wrap\_aw           & \scriptsize -0.265 \\
            \scriptsize basic\_nfunction\_types      & \scriptsize -0.261 \\
            \scriptsize basic\_ncontent\_types       & \scriptsize -0.253 \\
            \bottomrule
          \end{tabular}
        \end{subtable}\hfill
        \begin{subtable}
          [t]{0.237\textwidth}
          \centering
          \caption{simp -- sari}
          \begin{tabular}{lr}
            \toprule Variable            & $r$                        \\
            \midrule 
            \scriptsize root\_ttr\_aw  & \scriptsize 0.378 \\
            \scriptsize Word Count              & \scriptsize 0.378 \\
            \scriptsize nwords                  & \scriptsize 0.376 \\
            \scriptsize root\_ttr\_cw           & \scriptsize 0.365 \\
            \scriptsize basic\_ntypes           & \scriptsize 0.363 \\
            \scriptsize mtld\_ma\_wrap\_aw      & \scriptsize 0.357 \\
            \scriptsize basic\_ntokens          & \scriptsize 0.355 \\
            \scriptsize basic\_ncontent\_types  & \scriptsize 0.337 \\
            \scriptsize basic\_ncontent\_tokens & \scriptsize 0.327 \\
            \scriptsize mtld\_ma\_wrap\_cw      & \scriptsize 0.322 \\
            \bottomrule
          \end{tabular}
        \end{subtable}        
        \bigskip
        \begin{subtable}
          [t]{0.237\textwidth}
          \centering
          \caption{delta -- bertscore\_F1}
          \begin{tabular}{lr}
            \toprule Variable            & $r$                         \\
            \midrule 
            \scriptsize root\_ttr\_aw  & \scriptsize -0.518 \\
            \scriptsize basic\_ntypes           & \scriptsize -0.466 \\
            \scriptsize root\_ttr\_cw           & \scriptsize -0.464 \\
            \scriptsize nwords                  & \scriptsize -0.461 \\
            \scriptsize mtld\_ma\_wrap\_aw      & \scriptsize -0.457 \\
            \scriptsize Word Count              & \scriptsize -0.445 \\
            \scriptsize basic\_ncontent\_tokens & \scriptsize -0.434 \\
            \scriptsize basic\_ncontent\_types  & \scriptsize -0.432 \\
            \scriptsize basic\_ntokens          & \scriptsize -0.429 \\
            \scriptsize linsear                 & \scriptsize -0.413 \\
            \bottomrule
          \end{tabular}
        \end{subtable}\hfill
        \hspace{-1.4cm}
        \begin{subtable} 
          [t]{0.237\textwidth}
          \centering
          \caption{delta -- bleu}
          \begin{tabular}{lr}
            \toprule Variable                        & $r$                         \\
            \midrule 
            \scriptsize hyper\_verb\_noun\_Sav\_P1 & \scriptsize -0.403 \\
            \scriptsize hyper\_verb\_noun\_Sav\_Pav         & \scriptsize -0.384 \\
            \scriptsize hyper\_verb\_noun\_s1\_p1           & \scriptsize -0.362 \\
            \scriptsize hyper\_noun\_Sav\_P1                & \scriptsize -0.305 \\
            \scriptsize av\_pobj\_deps\_NN                  & \scriptsize -0.302 \\
            \scriptsize log\_ttr\_cw                        & \scriptsize -0.293 \\
            \scriptsize av\_pobj\_deps                      & \scriptsize -0.291 \\
            \scriptsize hyper\_noun\_S1\_P1                 & \scriptsize -0.291 \\
            \scriptsize KF\_Freq\_CW                        & \scriptsize 0.284  \\
            \scriptsize COCA\_fiction\_Freq\_CW        & \scriptsize 0.282  \\
            \bottomrule
          \end{tabular}
        \end{subtable}\hfill
          \hspace{-1cm}
        \begin{subtable}
          [t]{0.237\textwidth}
          \centering
          \caption{delta -- samsa}
          \begin{tabular}{lr}
            \toprule Variable                        & $r$                         \\
            \midrule 
            \scriptsize COCA\_news\_RL\_AW & \scriptsize 0.199  \\
            \scriptsize fog                                 & \scriptsize -0.198 \\
            \scriptsize COCA\_news\_RL\_CW          & \scriptsize 0.197  \\
            \scriptsize basic\_ncontent\_types              & \scriptsize -0.194 \\
            \scriptsize arindex                             & \scriptsize -0.193 \\
            \scriptsize basic\_ncontent\_tokens             & \scriptsize -0.192 \\
            \scriptsize COCA\_spoken\_RL\_AW        & \scriptsize 0.188  \\
            \scriptsize mtld\_ma\_wrap\_aw                  & \scriptsize -0.185 \\
            \scriptsize fkgl                                & \scriptsize -0.184 \\
            \scriptsize poly\_verb                          & \scriptsize 0.184  \\
            \bottomrule
          \end{tabular}
        \end{subtable}\hfill
        \hspace{-1.2cm}
        \begin{subtable}
          [t]{0.237\textwidth}
          \centering
          \caption{delta -- sari}
          \begin{tabular}{lr}
            \toprule Variable             & $r$                         \\
            \midrule 
            \scriptsize nwords          & \scriptsize -0.499 \\
            \scriptsize Word Count               & \scriptsize -0.496 \\
            \scriptsize root\_ttr\_aw            & \scriptsize -0.477 \\
            \scriptsize basic\_ntypes            & \scriptsize -0.461 \\
            \scriptsize basic\_ntokens           & \scriptsize -0.459 \\
            \scriptsize mtld\_ma\_wrap\_aw       & \scriptsize -0.449 \\
            \scriptsize basic\_nfunction\_types  & \scriptsize -0.447 \\
            \scriptsize basic\_nfunction\_tokens & \scriptsize -0.442 \\
            \scriptsize root\_ttr\_cw            & \scriptsize -0.420 \\
            \scriptsize basic\_ncontent\_tokens  & \scriptsize -0.410 \\
            \bottomrule
          \end{tabular}
        \end{subtable}
      \end{footnotesize}
      \caption{Top absolute values of significant correlation coefficients (\textit{p<0.5}) between human judgments and automatic metrics, on the SimplicityDA dataset.}
      \label{tab:simauto}
    \end{center}
              \end{footnotesize}

  \end{table*}

\begin{table*}[h!]
  \begin{footnotesize}
  \begin{center}
      \begin{tabular}{@{}c@{}}
        \noalign{\vskip 2mm}
                \begin{subtable}[t]{0.31\textwidth}
          \centering
          \caption{simp -- D-SARI}
          \begin{tabular}{lr}
            \toprule Variable & $r$ \\
            \midrule
            \scriptsize Word Count & \scriptsize -0.220 \\
            \scriptsize MRC\_Imageability\_FW & \scriptsize 0.172 \\
            \scriptsize Brysbaert\_CC\_FW & \scriptsize 0.171 \\
            \scriptsize MRC\_Concreteness\_FW & \scriptsize 0.164 \\
            \scriptsize MRC\_Meaningfulness\_FW & \scriptsize 0.148 \\
            \scriptsize COCA\_academic\_tri\_2\_DP & \scriptsize -0.146 \\
            \scriptsize KF\_Freq\_CW & \scriptsize 0.146 \\
            \scriptsize Kuperman\_AoA\_FW & \scriptsize -0.134 \\
            \scriptsize Brysbaert\_CC\_AW & \scriptsize 0.133 \\
            \scriptsize eat\_tokens & \scriptsize -0.129 \\
            \bottomrule
          \end{tabular}
        \end{subtable}\hfill                \begin{subtable}[t]{0.31\textwidth}
          \centering
          \caption{simp -- LENS}
          \begin{tabular}{lr}
            \toprule Variable & $r$ \\
            \midrule
            \scriptsize Word Count & \scriptsize -0.347 \\
            \scriptsize McD\_CD\_FW & \scriptsize 0.199 \\
            \scriptsize Kuperman\_AoA\_FW & \scriptsize -0.196 \\
            \scriptsize lsa\_average\_all\_cosine & \scriptsize 0.191 \\
            \scriptsize Brown\_Freq\_CW & \scriptsize 0.188 \\
            \scriptsize COCA\_magazine\_Range\_Log\_AW & \scriptsize -0.186 \\
            \scriptsize Brysbaert\_CC\_FW & \scriptsize 0.186 \\
            \scriptsize COCA\_academic\_tri\_2\_DP & \scriptsize -0.185 \\
            \scriptsize OG\_N\_H\_FW & \scriptsize 0.185 \\
            \scriptsize Freq\_N\_OGH\_FW & \scriptsize 0.185 \\
            \bottomrule
          \end{tabular}
        \end{subtable}\hfill               \begin{subtable}[t]{0.31\textwidth}
          \centering
          \caption{simp -- BERTScore\_F1}
          \begin{tabular}{lr}
            \toprule Variable & $r$ \\
            \midrule
            \scriptsize WN\_Mean\_Accuracy\_CW & \scriptsize -0.143 \\
            \scriptsize WN\_Mean\_Accuracy & \scriptsize -0.134 \\
            \scriptsize COCA\_Fiction\_Trigram\_Range\_Log & \scriptsize 0.133 \\
            \scriptsize LD\_Mean\_Accuracy\_CW & \scriptsize -0.131 \\
            \scriptsize COCA\_fiction\_tri\_2\_MI & \scriptsize -0.127 \\
            \scriptsize COCA\_fiction\_tri\_2\_MI2 & \scriptsize -0.112 \\
            \scriptsize LD\_Mean\_Accuracy & \scriptsize -0.112 \\
            \scriptsize COCA\_spoken\_Trigram\_Range\_Log & \scriptsize 0.096 \\
            \scriptsize LD\_Mean\_RT\_Zscore & \scriptsize 0.092 \\
            \scriptsize BNC\_Written\_Trigram\_Freq\_Normed\_Log & \scriptsize 0.090 \\
            \bottomrule
          \end{tabular}
        \end{subtable}
      \end{tabular}
      \vskip 3mm
      \begin{tabular}{@{}c@{}}
        \noalign{\vskip 2mm}
                \begin{subtable}[t]{0.31\textwidth}
          \centering
          \caption{delta -- D-SARI}
          \begin{tabular}{lr}
            \toprule Variable & $r$ \\
            \midrule
            \scriptsize WN\_Zscore\_CW & \scriptsize 0.305 \\
            \scriptsize COCA\_spoken\_tri\_MI2 & \scriptsize -0.303 \\
            \scriptsize WN\_Zscore & \scriptsize 0.298 \\
            \scriptsize COCA\_spoken\_tri\_MI & \scriptsize -0.283 \\
            \scriptsize COCA\_spoken\_Trigram\_Range\_Log & \scriptsize 0.266 \\
            \scriptsize WN\_Mean\_RT\_CW & \scriptsize 0.263 \\
            \scriptsize COCA\_spoken\_tri\_2\_MI2 & \scriptsize -0.258 \\
            \scriptsize Ortho\_N\_CW & \scriptsize -0.254 \\
            \scriptsize WN\_Mean\_RT & \scriptsize 0.252 \\
            \scriptsize PLD & \scriptsize 0.251 \\
            \bottomrule
          \end{tabular}
        \end{subtable}\hfill              \begin{subtable}[t]{0.31\textwidth}
          \centering
          \caption{delta -- LENS}
          \begin{tabular}{lr}
            \toprule Variable & $r$ \\
            \midrule
            \scriptsize COCA\_spoken\_tri\_2\_MI & \scriptsize -0.279 \\
            \scriptsize COCA\_spoken\_tri\_2\_MI2 & \scriptsize -0.267 \\
            \scriptsize LD\_Mean\_RT\_SD & \scriptsize 0.247 \\
            \scriptsize LD\_Mean\_RT\_SD\_CW & \scriptsize 0.241 \\
            \scriptsize COCA\_fiction\_Frequency\_AW & \scriptsize 0.241 \\
            \scriptsize COCA\_news\_Frequency\_AW & \scriptsize 0.235 \\
            \scriptsize COCA\_spoken\_tri\_MI2 & \scriptsize -0.234 \\
            \scriptsize COCA\_magazine\_Frequency\_AW & \scriptsize 0.226 \\
            \scriptsize Brown\_Freq\_CW\_Log & \scriptsize -0.221 \\
            \scriptsize poly\_noun & \scriptsize 0.217 \\
            \bottomrule
          \end{tabular}
        \end{subtable}\hfill
        \begin{subtable}[t]{0.31\textwidth}
          \centering
          \caption{delta -- BERTScore\_F1}
          \begin{tabular}{lr}
            \toprule Variable & $r$ \\
            \midrule
            \scriptsize AWL\_Sublist\_10\_Normed & \scriptsize 0.219 \\
            \scriptsize COCA\_academic\_tri\_T & \scriptsize -0.202 \\
            \scriptsize COCA\_magazine\_tri\_2\_T & \scriptsize -0.194 \\
            \scriptsize COCA\_academic\_tri\_2\_T & \scriptsize -0.192 \\
            \scriptsize COCA\_news\_tri\_T & \scriptsize -0.188 \\
            \scriptsize COCA\_magazine\_tri\_T & \scriptsize -0.186 \\
            \scriptsize COCA\_news\_tri\_2\_DP & \scriptsize 0.186 \\
            \scriptsize COCA\_news\_tri\_2\_T & \scriptsize -0.184 \\
            \scriptsize COCA\_Academic\_Trigram\_Frequency\_Log & \scriptsize -0.181 \\
            \scriptsize LD\_Mean\_RT\_SD\_CW & \scriptsize -0.179 \\
            \bottomrule
          \end{tabular}
        \end{subtable}
      \end{tabular}
      \vskip 3mm
    
  \end{center}
  \end{footnotesize}
  \caption{Top absolute values of significant correlation coefficients (p < .05) between human judgments and automatic metrics, on the D-WIKI dataset.}
  \label{tab:wikiauto}
\end{table*}

\normalsize

  \section{Experiments}

  \subsection{Readability Measures}
  First, we compute the correlations between the readability measures (metrics and
  features) themselves. Figures \ref{fig:heat_sent_diff} and \ref{fig:heat_sent_simp}
  show the correlation matrices computed on the SimplicityDA dataset (at the
  sentence level), respectively on the difference between the simplified and
  original sentences, and on the simplifications. Figures
  \ref{fig:heat_doc_diff} and \ref{fig:heat_doc_simp} show the correlation matrices computed on the D-Wikipedia dataset, respectively on the difference between the simplified and original sentences, and on the simplifications. We make three
  observations: (i) the measures mostly correlate with other measures of the
  same type, (ii) measures computed at the document-level show higher absolute values
  and (iii) measures computed on the difference between original texts and simplifications
  exhibit lower absolute values.

  \subsection{Measures and Human Judgment}

  \label{sec:read-judg} To compare readability measures (the features with the four
  readability tools, and the readability metrics) and human judgment, we compute
  them all on both datasets: SimplicityDA for the sentence-level (100 original
  sentences and 600 simplifications including 100 human-written ones) and D-Wikipedia
  for the document-level (100 original paragraphs and 500 simplifications
  including 100 human-written ones). For each dataset we compute the measures on
  both sides (original and simplified) separately. We compute the correlations with
  human judgment in two ways: (i) on the measures obtained on the
  simplifications only, and (ii) on the difference between the measures obtained
  on the original texts and the ones obtained on the simplifications. The first
  case focuses on simplicity, the second case focuses on simplification, by
  including a comparison with the original text.

  For both datasets, we report the correlations on the three criteria for human judgment:
  simplicity, fluency and meaning preservation.

  \subsection{Measures and Automatic Metrics}
  \label{sec:read-auto} To study the correlations between readability measures and
  automatic ATS metrics, we proceed in the same way as for the correlations
  between readability measures and human judgment. We report scores on the following
  automatic metrics: BLEU, SARI, BERTScore for simplicityDA (sentence-level), and BLEU, D-SARI, BERTScore, and LENS for D-Wikipedia (document-level).

  Regarding the metrics that require references (BLEU, SARI), for Simplicity-DA we use all the references
  that are provided, i.e. for each original sentence 10 references from ASSET \cite{alva-manchego-etal-2020-asset},
  1 from TurkCorpus \cite{xu-etal-2016-optimizing} and 1 from HSplit
  \cite{sulem-etal-2018-bleu}. For D-Wikipedia, we use the one reference
  simplification that is provided for each original text.

  \section{Results}
  \label{sec:results}

  \subsection{Measures and Human Judgment}
  \label{sec:read-judg-res}

  We report the top significant correlations between readability measures and human judgment at
  Table \ref{tab:simfeatures} for the SimplicityDA dataset, and Table \ref{tab:wikifeatures} for D-Wikipedia.

  For SimplicityDA, the highest absolute coefficient values are obtained with the meaning criterion computed on delta, with the top 10 ranging from -0.35 to -0.43. All of the other criteria have top absolute coefficient values between -0.15 and 0.30. In that regard, readability measures and human judgment on ATS at the sentence-level do not correlate well. We can observe all absolute values are higher when computed on the delta rather than on simplifications only. As the human judges were asked to rate simplification and not simplicity, this suggests that while the coefficient values are low, the difference between simplicity and simplification has an effect on both humans and measures.

  Regarding the D-Wikipedia dataset, the observations are similar: meaning exhibits the highest coefficient values, although with a higher discrepancy between the top 1 and 10 values (.433 vs .192 for simp-meaning and .342 vs .211 for delta-meaning). We found only 4 significant correlations for delta-grammar, which shows no sign of correlation with readability features in this set of observations (the highest value being .291 for delta, and .103 for simp). As for SimplicityDA, the values are generally higher for delta than for simp.

  Regarding simplicity, the values are low for both datasets. The most correlated set of observations is delta-simplicity with top 10 absolutevalues ranging from .234 to .298.

Not many features are found in more than one set of highest correlating values. For simplicityDA, we observe several kinds of type/token ratio (TTR): mostly root TTR and log TTR that are found in respectively 3 and 4 sets of observations out of 6. For DWiki, we see that the word count appears in 4 sets of observations, and corpus-based metrics (especially calibrated on COCA but also on the BNC) appear in 5 out of 6 sets.

  \subsection{Measures and Automatic Metrics}

  We report the correlations between readability measures and automatic metrics
  at Table \ref{tab:simauto} for the SimplicityDA dataset, and Table \ref{tab:wikiauto}
  for D-Wikipedia.

  For SimplicityDA, BERTScore has the highest correlation values, especially when the features are computed on delta (with a top 10 ranging from .413 to .518). SAMSA exhibits the lowest correlation values and is the only metric to correlate better when the features are computed on the simple texts only.

  Regarding D-Wikipedia, the correlations are generally lower. BERTScore has the lowest correlation values (from 0.09 to 0.219 across both computation modes), while LENS exhibits a slightly higher level of correlation than D-SARI.

  Regarding the features themselves, COCA-based featuers are present in all criteria with D-wiki, while they are only present for delta-samsa with SimplicityDA.

  TTR measures are present in 5 out of 8 sets of observations for SimplicityDA, and are completely absent for D-wiki. Those observations suggest that sentence simplification and document simplification evaluation do not entail the same phenomena. It appears quite surprising to see that TTR features correlate better with sentence-level simplification than with document-level simplification, as TTR it is frequently used for roughly assessing the complexity of a text, being a well-known measure of lexical diversity.

\section{Conclusion}

In this paper, we identified a knowledge gap in the literature on ATS, which is how readability measures correlate with human judgment and automatic metrics that are commonly used in the field. 
We consider it to be a relevant question, seeing that the ATS body literature shows uncertainty regarding correlations between human judgment and automatic metrics. Plus, as we have seen in Section \ref{sec:sota}, ATS studies start relying on readability measures in their methods or evaluation protocols, without theoretical grounding. We acknowledge that our findings do not go towards dissipating the uncertainty that the field has been experiencing. That said, we believe that our findings shed light onto a lack in the ATS ecosystem: a well-defined construct. 

\section*{Limitations}
The main limitation of our work is the amount and volume of data. We used the only data with human judgment that were available to us in English, so we could lead this study. These findings may vary on other corpora, other languages, and with other human annotators. While this impairs the generalizability of our study, we believe it reinforces our point that we should, as a community, focus more on clearly defining our task.

  \section{Bibliographical References}
  \label{sec:reference}

  \bibliographystyle{lrec2026-natbib}
  \bibliography{biblio}

  \section{Language Resource References}
  \label{lr:ref} \bibliographystylelanguageresource{lrec2026-natbib}
  \bibliographylanguageresource{languageresource}

\end{document}